\documentclass[11pt]{article}
\usepackage{emnlp2021}

\usepackage{times}
\usepackage{amsmath}
\usepackage{amsfonts}
\usepackage{latexsym}

\usepackage[linesnumbered,ruled,vlined]{algorithm2e}
\usepackage{graphicx} 
\usepackage{booktabs}
\usepackage{array}
\usepackage{enumitem}
\usepackage{amsthm}
\usepackage{algpseudocode}

\usepackage[switch]{lineno}
\usepackage[utf8]{inputenc}

\usepackage{eqparbox}
\usepackage[nopar]{lipsum}
\usepackage{multirow}
\usepackage{makecell}

\usepackage{xcolor}

\usepackage{scrextend}
\deffootnote[.25in]{.25in}{.15in}{\makebox[.25in][r]{\thefootnotemark .\hspace{.15in}}}

\makeatletter

\newtheorem{hypothesis}{Hypothesis}[section]
\newtheorem{theorem}{Theorem}[section]
\newtheorem{lemma}{Lemma}[section]

\newtheorem*{proof*}{Proof}

\usepackage{listings}
\usepackage{color}
\definecolor{codegreen}{rgb}{0.3,0.5,0.0}
\def\@fnsymbol#1{\ensuremath{\ifcase#1\or \dagger\or *\or \ddagger\or
   \mathsection\or \mathparagraph\or \|\or **\or \dagger\dagger
   \or \ddagger\ddagger \else\@ctrerr\fi}}

\title{Matching-oriented Product Quantization For Ad-hoc Retrieval}

\author{Shitao Xiao$^1$\thanks{Work is done during the internship at Microsoft.}, ~Zheng Liu$^2$\thanks{Corresponding author.}, ~Yingxia Shao$^1$\footnotemark[2], ~Defu Lian$^3$, ~Xing Xie$^2$\\
  1: Beijing University of Posts and Telecommunications, Beijing, China \\
  2: Microsoft Research Asia, Beijing, China \\
  3: University of Science and Technology of China, Hefei, China \\
  \texttt{\{stxiao,shaoyx\}@bupt.edu.cn} \\
  \texttt{\{zhengliu,xingx\}@microsoft.com} \\
  \texttt{defulian@ustc.edu.cn}}

\begin{document}
\maketitle

\begin{abstract}
Product quantization (PQ) is a widely used technique for ad-hoc retrieval. Recent studies propose supervised PQ, where the embedding and quantization models can be jointly trained with supervised learning. However, there is a lack of appropriate formulation of the joint training objective; thus, the improvements over previous non-supervised baselines are limited in reality. In this work, we propose the Matching-oriented Product Quantization (\textbf{MoPQ}), where a novel objective Multinoulli Contrastive Loss (\textbf{MCL}) is formulated. With the minimization of MCL, we are able to maximize the matching probability of query and ground-truth key, which contributes to the optimal retrieval accuracy. Given that the exact computation of MCL is intractable due to the demand of vast contrastive samples, we further propose the Differentiable Cross-device Sampling (\textbf{DCS}), which significantly augments the contrastive samples for precise approximation of MCL. We conduct extensive experimental studies on four real-world datasets, whose results verify the effectiveness of MoPQ. The  code  is  available  at https://github.com/microsoft/MoPQ.
\end{abstract}

\section{Introduction}
Ad-hoc retrieval is critical for many intelligent services, like web search and recommender systems. Given a query (e.g., a search request from user), ad-hoc retrieval selects relevant keys (e.g., webpages) from a massive set of candidates. It is usually treated as an approximate nearest neighbour search (ANNS) problem, where product quantization (PQ) \citep{jegou2010product} is one of the most popular solutions thanks to its competitive memory and time efficiency. PQ is built upon ``codebooks'', with which an input embedding can be quantized into a Cartesian product of ``codewords'' (preliminaries about the codebooks and codewords will be given in Section \ref{sec:2-1}). By this means, the original embedding can be compressed into a highly compact representation. More importantly, it significantly improves the retrieval efficiency, as query and key's similarity can be approximated based on the pre-computed distances between query and codewords. 

\textbf{Existing Limitation}. The original PQ algorithms  \citep{jegou2010product,ge2013optimized} are \textit{non-supervised}: based on the well-trained embeddings, the quantization model is learned with heuristic algorithms (e.g., $k$-means). In recent years, many works are dedicated to \textit{supervised PQ} \citep{cao2016deep,cao2017deep,klein2019end,gao2019beyond,chen2020differentiable}, where the embedding model and the quantization model are trained jointly. The supervised PQ requires an explicit training objective for the quantization model. Most of the time, ``the minimization of \textbf{reconstruction loss}'' is used for granted: the distortion between the original embedding ($\mathbf{z}$) and its quantization result ($\tilde{\mathbf{z}}$) needs to be reduced as much as possible for all the keys ($k$) in database: 
$min \sum\nolimits_{k} \| \mathbf{z}^k - \tilde{\mathbf{z}}^k  \|_2$. The above objective is seemingly plausible, as a ``small enough distortion'' will make the quantized embeddings equally expressive as the original embeddings. However, it implicitly hypothesizes that the distortion can be made sufficiently small, which is not always realistic in practice. This is because the reconstruction loss is subject to a lower-bound determined by the codebooks scale. As over-scaled codebooks result in prohibitive memory and time costs, there will always exist a positive reconstruction loss in reality. In this case, it can be proved that the minimization of reconstruction loss doesn't lead to the optimal retrieval accuracy. It's also empirically verified that the supervised PQ's advantage over the non-supervised baselines are limited and not consistently positive when the reconstruction loss minimization is taken as the training objective (see Section \ref{sec:2-2} and \ref{sec:exp} for detailed analysis).

\textbf{Our Work}. To address the above challenge, we propose the Matching-oriented Product Quantization (\textbf{MoPQ}), with a novel objective \textbf{MCL} formulated to optimize PQ's retrieval accuracy, together with a sample augmentation strategy \textbf{DCS} to ensure the effective minimization of MCL.

$\bullet$ The Multinoulli Contrastive Loss (MCL) is formulated as the new quantization training objective. The PQ-based ad-hoc retrieval can be probabilistically modeled by a cascaded generative process: 1) select codewords for the input key, based on which the quantized key embedding is composited; 2) sample query from the Multinoulli distribution determined by the quantized key embedding. The negative of the generation likelihood is referred as the Multinoulli Contrastive Loss (MCL). \textit{By minimizing MCL, the expected query-key matching probability will be optimized, which means the optimal retrieval accuracy}.

$\bullet$ The contrastive sample augmentation is designed to facilitate the minimization of MCL. The computation of MCL is intractable as it requires the normalization over vast contrastive samples (the quantized embeddings of all the keys). Thus, it has to be approximated by sampling whose effect is severely affected by sample size. In this work, we propose the Differentiable Cross-device Sampling (DCS), {where a distributed embedding sharing mechanism is introduced for contrastive sample augmentation.} As the gradients are stopped at the cross-device shared embeddings, we propose a technique called the ``combination of Primary and Image Losses'', where the shared embeddings are made virtually differentiable to keep the model update free from distortions.

In summary, our work identifies a long-existing defect about the training objective of supervised PQ; meanwhile, a simple but effective remedy is proposed, which optimizes the expected retrieval accuracy of PQ. We make extensive experimental studies with four benchmark text retrieval datasets, where our proposed methods significantly outperform the SOTA supervised PQ baselines. Our code and datasets will be made public-available to facilitate the research progress in related areas.

\section{Revisit of Supervised PQ}\label{sec:2}
We start with the preliminaries of PQ's application in ad-hoc retrieval. Then, we analyze the defect of taking the reconstruction loss minimization as supervised PQ's training objective. 

\subsection{Preliminaries}\label{sec:2-1}

$\bullet$ \textbf{Product Quantization (PQ)}. PQ is a popular approach for learning quantized representations. It is based on the foundation of $M$ codebooks $\mathbf{C}$: $\{\mathbf{C}_1, ... , \mathbf{C}_M\}$; each $\mathbf{C}_i$ consists of $L$ codewords: $\{\mathbf{C}_{i1}, ... , \mathbf{C}_{iL}\}$; each $\mathbf{C}_{ij}$ is a vector whose dimension is $d/M$. Given an embedding $\mathbf{z}$ (with dimension $d$), it is firstly sliced into to $M$ sub-vectors [$\mathbf{z}_1, \mathbf{z}_2,..., \mathbf{z}_M$]; then the sub-vector $\mathbf{z}_i$ is assigned to one of the codewords of codebook $\mathbf{C}_i$, whose ID is denoted by the one-hot vector $\textbf{b}_i$. 
The assignment is made by the codeword selection function, which maps each sub-vector $\mathbf{z}_i$ to its most relevant codeword, e.g., $\mathbf{b}_i=one\_hot(argmin{||\mathbf{z}_i-\mathbf{C}_{i*}||_2})$.
As a result, the embedding $\mathbf{z}$ is quantized into a collection of codes: $\textbf{B} = \{\textbf{b}_1,...,\textbf{b}_M\}$, where the embedding itself is approximated by the concatenation of the assigned codewords: $\tilde{\mathbf{z}} = [\mathbf{C}_1\textbf{b}_1,...,\mathbf{C}_M\textbf{b}_M]$.

Non-supervised PQ takes the well-trained embeddings as input, and learns the quantization model with heuristic algorithms, like $k$-means. In contrast, supervised PQ jointly learns the embedding and quantization models based on labeled data (the paired query and key).
Specifically, it learns the query and key's embeddings: $\mathbf{z}^q$ and $\mathbf{z}^k$, such that the query and key's relationship can be predicted based on their inner product $\langle\mathbf{z}^q, \mathbf{z}^k\rangle$. Furthermore, it learns the codebooks where the quantized embeddings may preserve the same expressiveness as the original embeddings.

$\bullet$ \textbf{PQ for ad-hoc retrieval}. PQ is also widely applied for ad-hoc retrieval. On one hand, the float vectors are quantized for high memory efficency. On the other hand, the retrieval process can be significantly accelerated. For each key within database, the embedding $\mathbf{z}^k$ is quantized as $\tilde{\mathbf{z}}_k=[\mathbf{C}_1\textbf{b}_1,...,\mathbf{C}_M\textbf{b}_M]$\footnote{\scriptsize We adopt the Asymmetric Distance Computation (ADC) \citep{jegou2011searching}, where only keys need to be quantized.}. For a given query embedding $\mathbf{z}^q$, the inner-product with the $M$ codebooks can be enumerated and kept within the distance table $\textbf{T}_q$, where $\textbf{T}_q[i,j]=\langle\mathbf{z}^q_i, \mathbf{C}_{ij}\rangle$. Finally, the query and key's inner product $\langle\mathbf{z}^q, \tilde{\mathbf{z}}^k\rangle$ can be efficiently derived by looking up the pre-computed results within $\textbf{T}_q$: $\sum_{1,...,M}\textbf{T}_q[i,\textbf{b}_i]$, where no more dot product computations are needed.

\subsection{Defect of Reconstruction Loss}\label{sec:2-2}
The reconstruction loss minimization is usually adopted as the quantization model's training objective in supervised PQ \citep{cao2016deep,gao2019beyond,chen2020differentiable}. It requires the distortions between key embedding and its quantization result to be reduced as much as possible:
\begin{equation}\label{eq:reconstruct}
    min \sum\nolimits_k \| \mathbf{z}^k - \tilde{\mathbf{z}}^k \|_2.
\end{equation}
The minimization of reconstruction loss is seemingly plausible given the following hypothesises.

\begin{hypothesis}\label{hp:1}
The quantized keys' embeddings are less accurate in predicting query and key's relevance, compared with the non-quantized ones.
\end{hypothesis}

\begin{hypothesis}\label{hp:2}
The loss of prediction accuracy is a monotonously increasing function of the reconstruction loss (defined in Eq. \ref{eq:reconstruct}).
\end{hypothesis}

The first hypothesis can be ``assumed correct'' in reality, considering that the quantized embeddings are less expressive than the original embeddings (due to the finite number of codewords). However, the second hypothesis is problematic. In the following part, we analyze the underlying defect from both theoretical and empirical perspectives. 

\subsubsection{Theoretical Analysis\footnote{\scriptsize Proofs of Theorem~\ref{thm:1} and Lemma~\ref{lemma:1} are put to the supplementary material.}}
We theoretically analyze two properties about the reconstruction loss: 1) it is indelible; and 2) decreasing of the reconstruction loss does not necessarily improve the prediction accuracy.

\begin{theorem}
\label{thm:1}
\emph{(\textbf{Positive Reconstruction Loss})} 
The reconstruction loss is positive if the codebooks' scale is smaller than the key's scale. 
\end{theorem}
That is to say, the input will always be changed after quantization given a reasonable scale of codebooks, making it impossible to keep the quantized embeddings equally expressive as the original embeddings by eliminating the reconstruction loss. Moreover, we further show that the reduction of the reconstruction loss doesn't necessarily improve the prediction accuracy.



\begin{figure*}[t]
\centering
\includegraphics[width=1.0\textwidth]{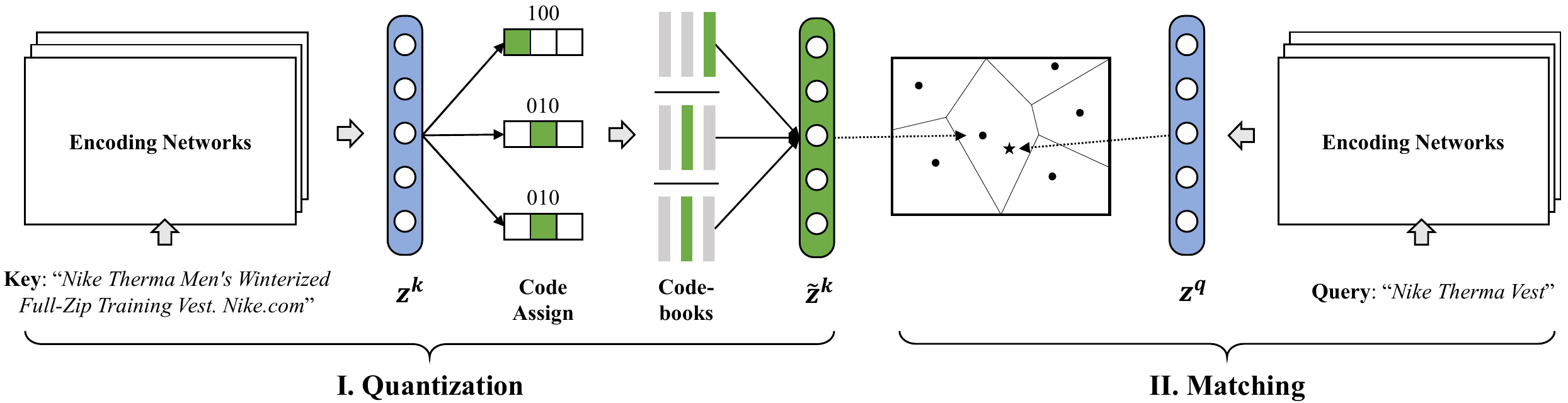}
\caption{\small PQ's retrieval workflow: (I) Quantization: the key's embedding ($\mathbf{z}^k$) is assigned to codes, whose related codewords are composited as the quantized key embedding ($\tilde{\mathbf{z}}^k$); (II) Matching: the quantized key embedding (one of the centroids of the Voronoi diagram) confines the targeted query embedding $\mathbf{z}^q$ within its own Voronoi cell.}
\label{fig:1}
\end{figure*}

\renewcommand{\arraystretch}{1.35}
\newcolumntype{C}[1]{>{\centering\let\newline\\\arraybackslash\hspace{0pt}}m{#1}}
\newcommand\ChangeRT[1]{\noalign{\hrule height #1}}

\begin{table*}[t]
    \centering
    \scriptsize
    \begin{tabular}{p{1.2cm} | C{1.2cm} C{1.2cm} | C{1.2cm} C{1.2cm} |  C{1.2cm} C{1.2cm} | C{1.2cm} C{1.2cm} }
    \ChangeRT{1pt}
    & \multicolumn{2}{c}{\textbf{Search Ads}} & \multicolumn{2}{|c}{\textbf{Quora}} & \multicolumn{2}{|c}{\textbf{News}} & \multicolumn{2}{|c}{\textbf{Wiki}} \\
    \cmidrule(lr){1-1}
    \cmidrule(lr){2-3}
    \cmidrule(lr){4-5}
    \cmidrule(lr){6-7}
    \cmidrule(lr){8-9}
        \textbf{Methods} & 
        \textbf{R-loss} & \textbf{Recall} & \textbf{R-loss} & \textbf{Recall} & \textbf{R-loss} & \textbf{Recall} & \textbf{R-loss} & \textbf{Recall} \\
        \hline
        {DQN 1e-3} & 16.6820 & 0.1045 & 16.6647 & 0.5172 & 20.2182 & 0.2357 & 18.3501 & 0.0535  \\
        \hline
        {DQN 1e-2} & 7.5604 & \textbf{0.1433} & 5.6794 & \textbf{0.5436} & 9.2987 & \textbf{0.3138} & 6.9921 & \textbf{0.0720} \\
        \hline
        {DQN 1e-1} & 1.4021 & 0.1132 & 2.5604 & 0.4798 & 1.9051 & 0.2636 & 1.4542 & 0.0547 \\
        \hline
        {DQN 1.0}  & \underline{0.2575} & 0.0575 & \underline{0.5038} & 0.1869 & \underline{0.7624} & 0.1488 & \underline{0.3043} & 0.0085 \\
    \ChangeRT{1pt}
    \end{tabular}
    \caption{\small Relationships between the reconstructive loss (R-Loss) and PQ's accuracy (Recall@10). The reconstruction loss is monotonously reduced when its weight becomes larger. However, the smallest reconstruction loss (marked by ``$\_\_$'') doesn't lead to the optimal accuracy (marked in bold).}
    \label{tab:analysis}
\end{table*}

We start by showing the existence of ``{quantization invariant perturbation}'', i.e., the codeword assignment will stay unchanged when such perturbations are added to the codebooks.

\begin{lemma}\label{lemma:1}
For each codebook $\mathbf{C}_i$, there will always exist perturbation vectors like $\epsilon_{i}$, where the manipulation of codewords: $\hat{\mathbf{C}}_{i*} \leftarrow \mathbf{C}_{i*}+\epsilon_{i}$, doesn't affect the codeword assignment.  
\end{lemma}

On top of the existence of quantization invariant perturbations, we may derive the ``Non-monotone'' about the relationship between the prediction accuracy and the reconstruction loss.
\begin{theorem}
\emph{(\textbf{Non-Monotone})} PQ's prediction accuracy is not monotonically increasing with the reduction of the reconstruction loss.
\end{theorem}
\begin{proof*}
The statement is proved by contradiction. 
For each codebook, we generate a perturbation which satisfies Lemma~\ref{lemma:1} and add it to the codewords: $\hat{\mathbf{C}}_{i*} \leftarrow \mathbf{C}_{i*} + \epsilon_i$. According to the Lemma~\ref{lemma:1}, the codeword assignment will not change, so the quantized key embedding become: $\hat{\mathbf{z}}^{k} = \tilde{\mathbf{z}}^k + \epsilon$, where $\epsilon=[\epsilon_1, \epsilon_2, ..., \epsilon_M]$. Now we may derive the following relationship about the reconstruction losses:
\begin{align*}
    \mathbb{E} \|\mathbf{z}^k-\hat{\mathbf{z}}^{k}\|_2 &= 
    \big(
    \mathbb{E} (\mathbf{z}^k-\tilde{\mathbf{z}}^{k} )^2 + \mathbb{E} (\epsilon)^2
    \big)^{1/2}  \\ &> 
    \mathbb{E} \|\mathbf{z}^k-\tilde{\mathbf{z}}^k\|_2.
\end{align*}
(The first equivalence holds conditioned on the independence between $\mathbf{z}^k-\tilde{\mathbf{z}}^{k}$ and 
$\epsilon$.)\footnote{\scriptsize The $\epsilon$ is generated based on an arbitrary unit vector, so the independence condition always holds.}
That is to say, the reconstruction loss is increased after the perturbation. At the same time, we may also derive the query and key's relationship before ($\mathrm{R}$) and after ($\hat{\mathrm{R}}$) the perturbation:
\begin{equation*}
\begin{split}
    &\hat{\mathrm{R}}(q,k) 
    =
    \frac{\exp\big(\langle\mathbf{z}^q,\hat{\mathbf{z}}^{k}\rangle\big)}{\sum\nolimits_{k'} \exp\big(\langle\mathbf{z}^q,\hat{\mathbf{z}}^{k'}\rangle\big)} 
    \\
    &=
    \frac{\exp\big(\langle\mathbf{z}^q,\tilde{\mathbf{z}}^k+\epsilon\rangle\big)}{\sum\nolimits_{k'} \exp\big(\langle\mathbf{z}^q,\tilde{\mathbf{z}}^{k'}+\epsilon\rangle\big)} 
     \\
    &=
     \frac{\exp\big(\langle\mathbf{z}^q,\tilde{\mathbf{z}}^k\rangle\big)}{\sum\nolimits_{k'} \exp\big(\langle\mathbf{z}^q,\tilde{\mathbf{z}}^{k'}\rangle\big)}  
    =
    \mathrm{R}(q,k).
\end{split}
\end{equation*}
In other words, the relationship between query and key is preserved despite the increased reconstruction loss. Thus, a contradiction is obtained for the monotonous relationship between the prediction accuracy and reconstruction loss. $\square$
\end{proof*}

\subsubsection{Empirical Analysis}
To further verify the theoretical conclusions, we empirically analyze the relationship between the prediction accuracy and the reconstruction loss as Table \ref{tab:analysis}, by taking the deep quantization network \citep{cao2016deep} (DQN for short) as the example\footnote{\scriptsize More detailed experiment settings are given in Section \ref{sec:exp}.}. The weight of reconstruction loss is tuned as: 1, 1e-1, 1e-2, 1e-3 (larger weights will lead to higher loss reduction), and the weight of embedding learning is fixed to 1. We find that the reconstruction loss is monotonously reduced when a larger learning weight is used. However, the smallest reconstruction loss doesn't bring the highest prediction accuracy (measured with Recall@10), which echos our theoretical analysis. More experimental studies in Section \ref{sec:exp} demonstrate that the supervised PQ's advantages over the non-supervised baselines are inconsistent and sometimes insignificant, when the reconstruction loss minimization is taken as the objective (even with the optimally tuned weights).

To summarize, the reconstruction loss cannot be eliminated with a feasible scale of codebooks, and the reduction of the reconstruction loss doesn't necessarily improve the retrieval accuracy. As such, we turn around to formulate a new objective, where the model will stay with the reconstruction loss but optimize the PQ's retrieval accuracy.


\section{Matching-oriented PQ}
We present the Matching-oriented PQ (MoPQ) in this section. In MoPQ, a new quantization objective MCL is proposed, whose minimization optimizes the query-key's matching probability, therefore bringing about the optimal retrieval accuracy. Besides, the DCS method is introduced, which facilitates the effective minimization of MCL.

\subsection{Multinoulli Contrastive Loss}\label{sec:pq-mlc}
\begin{figure}[t]
\centering
\includegraphics[width=0.67\linewidth]{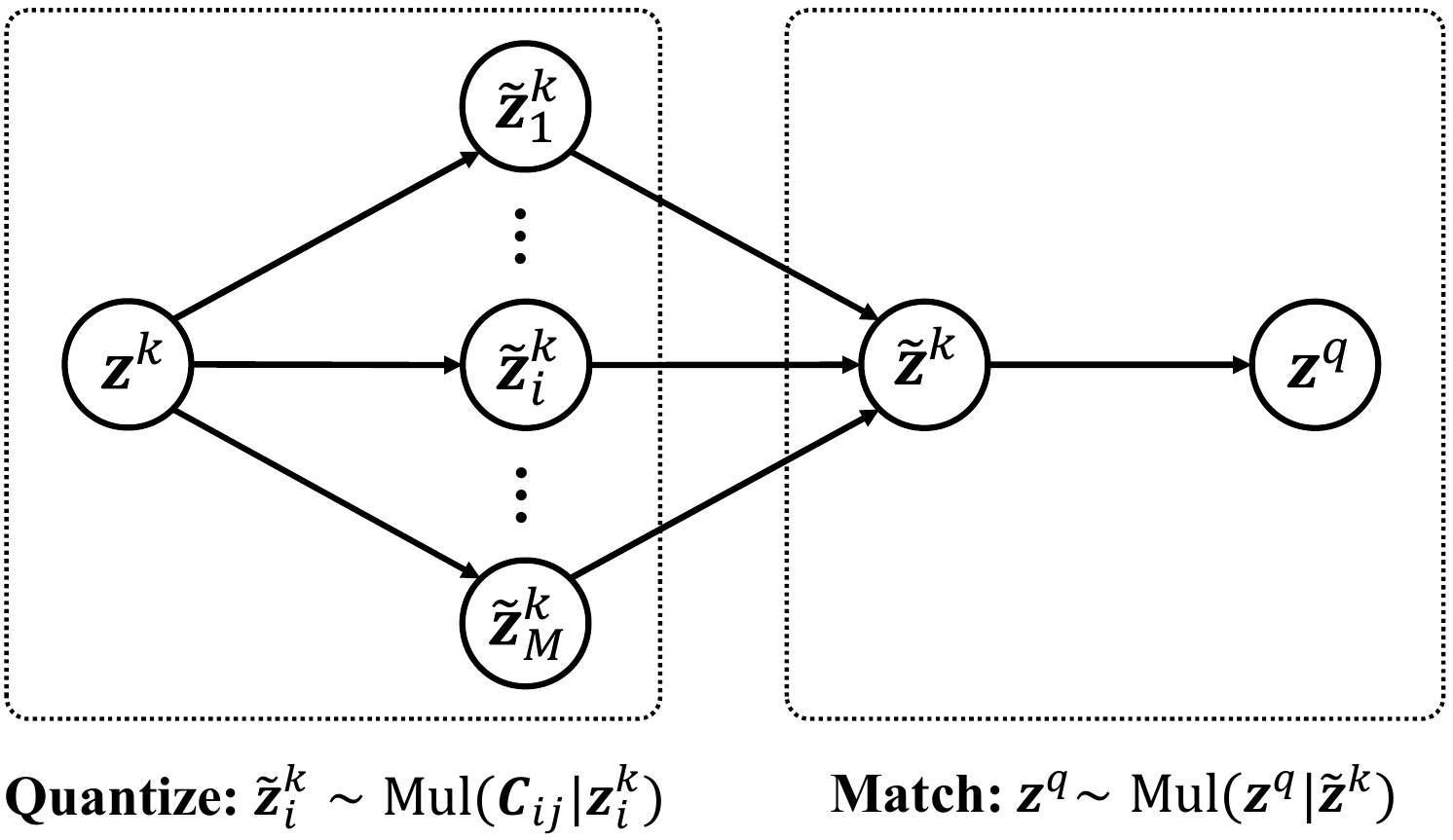}
\caption{\small The Multinoulli Generative Process.}
\label{fig:graph}
\end{figure}

The ad-hoc retrieval with PQ can be divided into two stages, shown as Figure \ref{fig:1}. The first stage is called Quantization. The key embedding is assigned to a series of binary codes, each of which is corresponding to one codeword within a codebook. The assigned codewords are composited as the quantized key embedding, which is a centroid of the Voronoi diagram determined by the codebooks. The second stage is called Matching, where the quantized key embedding confines the targeted query embedding within its own Voronoi cell. The above matching process is probabilistically modeled as the \textbf{Multinoulli Generative Process} (as Figure \ref{fig:graph}):
\begin{itemize}[leftmargin=10pt]
    \item For each codebook $i$, a codeword is sampled from the Multinoulli distribution: $\mathrm{Mul}(\mathbf{C}_{ij}|\mathbf{z}^k_i)$, denoted as $\tilde{\mathbf{z}}^k_i$; the quantized key embedding $\tilde{\mathbf{z}}^k$ is generated as the concatenation of $\{\tilde{\mathbf{z}}^k_i\}_M$ .
    \item The query is sampled from the distribution: $\mathrm{Mul}(\mathbf{z}^q|\tilde{\mathbf{z}}^k)$, parameterized by the quantized key embedding $\tilde{\mathbf{z}}^k$ and query embedding $\mathbf{z}_q$.
\end{itemize}
Thus, the matching probability can be factorized by the joint distribution of making codeword selection from all codebooks: $\prod\nolimits_{i} P(\mathbf{C}_{ij}|\mathbf{z}^k_i)$, and sampling query based on $\mathbf{z}^q$ and $\tilde{\mathbf{z}}^k$: $P(\mathbf{z}^q|\tilde{\mathbf{z}}^k)$:
\begin{equation*}
P(\mathbf{z}^q|\mathbf{z}^k) = \sum\nolimits_{j}\prod\nolimits_{i} P(\mathbf{C}_{ij}|\mathbf{z}^k_i)P(\mathbf{z}^q|\tilde{\mathbf{z}}^k).
\end{equation*}
(``$\sum\nolimits_{j}$'' indicates the enumeration of all possible codeword selection.) We expect to maximize the above query-key matching probability so as to achieve the optimal retrieval accuracy. However, the exact calculation is intractable due to: 1) the almost infinite combinations of codewords, and 2) the unknown distribution of the queries. In this place, the following transformations are made.

Firstly, we leverage the Straight Through Estimator \cite{bengio2013estimating}, where $P(\mathbf{C}_{ij}|\mathbf{z}^k_i)$ is transformed by the hard thresholding function:
\begin{equation*}
    P'(\mathbf{C}_{ij}|\mathbf{z}^k_i) = 
    \begin{cases}
    1, ~\text{ if }~ j=argmax\{P(\mathbf{C}_{i*}|\mathbf{z}^k_i)\}; \\
    0, ~\text{ otherwise.}
    \end{cases}
\end{equation*}
The above probability is calculated as $P' = (P'-P).sg() + P$ such that the gradients can be back propagated (``$sg()$ '' is the stop gradient operation). Now the generation probability is simplified as:
\begin{equation*}
P(\mathbf{z}^q|\mathbf{z}^k) = \prod\nolimits_{i} P'(\mathbf{C}_{ij^*}|\mathbf{z}^k_i) P(\mathbf{z}^q|\tilde{\mathbf{z}}^k),
\end{equation*}
where $j^* = argmax\{P(\mathbf{C}_{i*}|\mathbf{z}^k_i)\}$. (``$\sum_{j}$'' can be removed because $P'(\mathbf{C}_{ij}|\mathbf{z}^k_i)=0$, $\forall j{\neq}j^*$.) 

Secondly, we make a further transformation for the query's generation probability:
\begin{equation*}
    P(\mathbf{z}^q|\tilde{\mathbf{z}}^k) = \frac{P(\tilde{\mathbf{z}}^k|\mathbf{z}^q)P(\mathbf{z}^q)}{P(\tilde{\mathbf{z}}^k) } \propto P(\tilde{\mathbf{z}}^k|\mathbf{z}^q),
\end{equation*}
where $P(\mathbf{z}^q)$ and $P(\tilde{\mathbf{z}}^k)$ are the prior probabilities regarded as unknown constants. The conditional probability $P(\tilde{\mathbf{z}}^k|\mathbf{z}^q)$ calls for the normalization over the quantized key embeddings $\{\tilde{\mathbf{z}}^k\}$, which is deterministic and predefined. Now the generation probability is transformed as:
\begin{gather}\label{eq:3-1}
P(\mathbf{z}^q|\mathbf{z}^k) \propto \prod\nolimits_{i} P'(\mathbf{C}_{ij^*}|\mathbf{z}^k_i) P(\tilde{\mathbf{z}}^k|\mathbf{z}^q) = 
\\
\prod_i\frac{\exp(s(\mathbf{z}^k_i,\mathbf{C}_{ij}))}{\sum_{j} \exp(s(\mathbf{z}^k_i,\mathbf{C}_{ij}))} * \frac{\exp(\langle\tilde{\mathbf{z}}^k,\mathbf{z}^q\rangle)}{\sum_{k'}\exp(\langle \tilde{\mathbf{z}}_{k'},\mathbf{z}^q\rangle)}, \nonumber
\end{gather}
where ``$s(\cdot)$'' is the code assignment function (detailed forms will be discussed in Section \ref{sec:exp-1}).

The negative logarithm of the final simplification in Eq. \ref{eq:3-1} is called the \textbf{Multinoulli Contrastive Loss} (MCL). It is used as our quantization training objective, whose minimization optimizes the query and key's matching probability.

\begin{figure*}[t]
\centering
\includegraphics[width=0.80\textwidth]{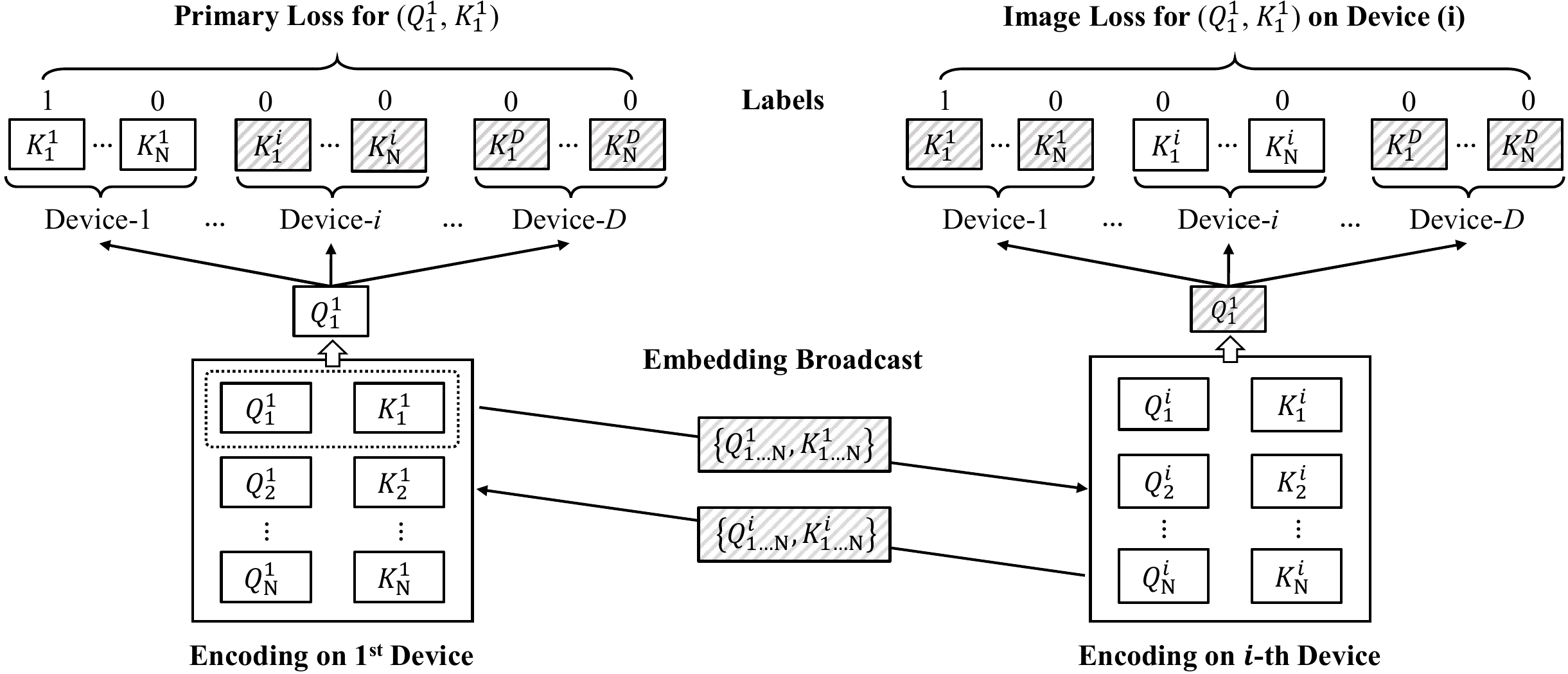}
\caption{\small Contrastive sample augmentation based on DCS. The primary loss and the image loss are combined to make the contrastive samples shared from other devices (shaded boxes) virtually differentiable.}
\label{fig:2}
\end{figure*}

\setlength{\textfloatsep}{10pt}
\begin{algorithm}[t]
\caption{DCS Method}\label{alg:1}
    \LinesNumbered 
    \Begin{
        \For{Device $i=$ 1 ... D}{
        $\{Q^i_*, K^i_*\}_N \leftarrow $ Encode($\{q^i_*,k^i_*\}_N$)\;  
        broadcast $\{Q^i_*, K^i_*\}_N$\;
        } 
        \For{Device $i=$ 1 ... D}{
        $L^i_p \leftarrow$ primary loss\;
        \For{Device $j{\neq}i$}{
        $L^{ji}_c \leftarrow$ image loss for Device-$j$\;
        }
        $\nabla_{\boldsymbol{\theta}}^i \leftarrow \nabla_{\boldsymbol{\theta}}\big(L^i_p + \sum_j L^{ji}_c\big)$
        }
        Update model w.r.t. $\sum_i\nabla_{\boldsymbol{\theta}}^i$.
    }
\end{algorithm}

\subsection{Approximating MCL with DCS}\label{sec:pq-DCS}
MCL calls for the normalization over all keys' quantized embeddings $\{\tilde{\mathbf{z}}^k\}$, whose computation cost is huge. It has to be approximated by negative sampling \citep{bengio2013estimating, facebook_retrival}, where a subset of keys are encoded as the normalization term. Recent works \citep{gillick-etal-2019-dense_learning,luan2020sparse,wu-etal-2020-entity_retrival,karpukhin-etal-2020-OQA} use in-batch contrastive samples, which are free of extra encoding cost: for the $i$-th training instance within a mini-batch, the $j$-th key's quantized embedding ($j \neq i$) will be used as a contrastive sample. Thus, there will be $N-1$ cost free contrastive samples in total ($N$: batch size). Recent studies also leverage cross device in-batch sampling for sample augmentation in distributed environments \citep{ding2020rocketqa}. Particularly, a training instance on one device may take advantage of quantized key embeddings on other devices as its contrastive samples. Thus, the contrastive samples will be increased by $\times{D}$ times (${D}$: the number of devices). A problem about the cross device in-batch sampling is that the shared embeddings from other devices are not differentiable (because the shared values need to be detached from their original computation graphs). As a result, the partial gradients cannot be back-propagated through the cross-device contrastive samples, which causes distortions of the model's update, thus undermines the optimization effect.

$\bullet$ \textbf{DCS Method}. We propose the \textbf{D}ifferentiable \textbf{C}ross-device in-batch \textbf{S}ampling (\textbf{DCS}), which enables partial gradients to be back propagated for the cross-device contrastive samples. The overall workflow is summarized with Alg. \ref{alg:1}, where the core technique is referred as the ``combination of \textbf{Primary} and \textbf{Image} losses''. Suppose a total of $D$ GPU devices are deployed, each one processes $N$ training instances. The training instances encoded on the $i$-th Device are denoted as $\{Q^i_{1...N},K^i_{1...N}\}$.\footnote{\scriptsize $Q^i_j$ and $K^i_j$: the query embedding ($\mathbf{z}^q$) and the quantized key embedding ($\mathbf{\tilde{z}}^k$) of the $j$-th training instance on device $i$.} The embeddings generated on each device will be broadcasted to all the other devices. As a result, the $i$-th device will maintain two sets of embeddings: 1) $\{Q^i_{1...N},K^i_{1...N}\}$, which are locally encoded and differentiable, and 2) $\{Q^{{\neq i}}_{1...N},K^{{\neq i}}_{1...N}\}$, which are broadcasted from other devices and thus non-differentiable. Each training instance $(Q^i_j,K^i_j)$ will have two losses computed in parallel: the primary loss on the device where it is encoded (i.e., Device-$i$), and the image loss on the devices where it is broadcasted. 

Take the first instance on Device-1 ($Q^1_1,K^1_1$) for illustration (as Figure \ref{fig:2}). The query-key matching probability $P(K^1_1|Q^1_1)$ on Device-1 is:
\begin{equation}\label{eq:6}
    \frac{\exp({\langle Q_1^1, K_1^1 \rangle})}{\sum_j  \exp({\langle Q_1^1, K_j^1 \rangle}) 
    + {\scriptstyle\sum}_{{\neq}1} \exp({\langle Q_1^1, \overline{K_j^*} \rangle})}.
\end{equation}
The cross-device embeddings are detached, therefore, the partial gradients are stopped at these variables (marked as $\overline{Q}$ and $\overline{K}$). The above query-key matching probability will be used by MCL (for ``$P(\mathbf{\tilde{z}_k}|\mathbf{z_q})$'') in Eq. \ref{eq:3-1}, whose result is called the \textbf{primary loss} w.r.t. ($Q^1_1,K^1_1$); the sum of primary losses for all ($Q^1_*,K^1_*$) is denoted as $L_p^{1}$.

Meanwhile, the query-key matching probability $P(K^1_1|Q^1_1)$ is also computed on all other devices. For the $i$-th device ($i\neq1$), $P(K^1_1|Q^1_1)$ becomes:
\begin{equation}\label{eq:7}
    \frac{\exp({\langle \overline{Q_1^1}, \overline{K_1^1} \rangle})}{\sum_j  \exp({\langle\overline{ Q_1^1}, K_j^i \rangle}) 
    + {\scriptstyle\sum}_{{\neq}i} \exp({\langle \overline{Q_1^1}, \overline{K_j^*} \rangle})}.
\end{equation}
The differentiability is partially inverted compared with $P(K^1_1|Q^1_1)$ in Eq. \ref{eq:6}: $K_j^i$ becomes differentiable, but $Q_1^1$ and $K_j^1$ become non-differentiable. The above probability is used to derive another MCL, which is called the \textbf{image loss} of ($Q^1_1,K^1_1$) on Device-$i$; the sum of image losses of all ($Q^1_*,K^1_*$) on Device-$i$ is denoted as $L_c^{1i}$. Clearly, the above image loss will compensate the stopped gradients (related to $K^i_j$) in the primary loss.

The primary losses and image losses are gathered from all GPU devices and added up, based on which the model parameters $\boldsymbol{\theta}$ are updated w.r.t. the following partial gradients:
\begin{equation}\label{eq:8}
    \nabla_{\boldsymbol{\theta}} \Big(\sum\nolimits_i \big( L_p^{i} + {\sum\nolimits}_{j{\neq}i}L_c^{ij} \big)\Big).
\end{equation}
It can be verified that the above results are equivalent to the partial gradients derived from the following full-differentiable distributions:
\begin{equation}\label{eq:9}
    \sum_{k=1}^{N}\sum_{l=1}^{N}\frac{\exp(\langle Q_l^k, K_l^k \rangle)}
    {\sum_{j=1}^N\sum_{i=1}^D \exp(\langle Q_k^1, K_j^i \rangle) }.
\end{equation}
Thus, the partial gradients are free from distortions caused by the non-differentiable variables, which enables MCL to be precisely approximated with the cross-device augmented contrastive samples.

\section{Experimental Studies}\label{sec:exp}

\subsection{Experiment Settings}\label{sec:exp-1}
$\bullet$ \textbf{Data}. We use three open datasets. \textbf{Quora}\footnote{\scriptsize https://www.kaggle.com/c/quora-question-pairs}, with question pairs of duplicated meanings~\cite{wang2019kepler}; we use one question to retrieve its counterpart. \textbf{News}\footnote{\scriptsize https://msnews.github.io}, with news articles from Microsoft News~\cite{mind}; we use the headline to retrieve the news body. \textbf{Wiki}\footnote{\scriptsize https://deepgraphlearning.github.io/project/wikidata5m}, with passages from Wikipedia; we use the first sentence to retrieve the remaining part of the passage. One large industrial dataset \textbf{Search Ads}, with user's search queries and clicked ads from a worldwide search engine; we use search queries to retrieve the titles of clicked ads. (As Table \ref{tab:dataset}.)

$\bullet$ \textbf{Baselines}. We consider both supervised and non-supervision PQ as our baselines. For supervised PQ, the embedding model and the quantization model are learned jointly. For non-supervised PQ, the embedding model is learned at first; then the quantization model is learned with fixed embeddings. We consider the following supervised methods. \textbf{DQN} \citep{cao2016deep}, which is learned with two objectives: the embedding model is learned to match the query and key, and the quantization model is learned to minimize the reconstruction loss. \textbf{DVSQ} \citep{cao2017deep}, which adapts the reconstruction loss to minimize the distortion of query and key's inner product. \textbf{SPQ} \citep{klein2019end}, which minimizes the disagreement between the hard and soft allocated codewords. \textbf{DPQ} \citep{chen2020differentiable}, which still minimizes the reconstruction loss as DQN, but leverages a different quantization module. We also include 2 non-supervised baselines: the vanilla \textbf{PQ} \cite{jegou2011searching}, and \textbf{OPQ} \citep{ge2013optimized}.Although OPQ is non-supervised, it learns the transformation of the input embeddings such that the reconstruction loss can be minimized.

\begin{table}[t]
    \centering
    \scriptsize
    \begin{tabular}{p{1.0cm} p{1.0cm} p{1.0cm} p{1.0cm} p{1.0cm}}
    \ChangeRT{1pt}
        & \textbf{Train} & \textbf{Valid} & \textbf{Test} & \textbf{\#Keys}\\
    \hline
        News & 79,122 & 9,891 & 9,891 & 98,388 \\
        Quora & 68,240 & 9,055 & 24,041 & 537,340 \\
        Wiki & 289,623 & 10,000 & 10,000 & 1,902,625 \\
        Ads & 1,169,453 & 10,000 & 10,000 & 1,738,237 \\
    \ChangeRT{1pt}
    \end{tabular}
    \caption{\small Specifications of Datasets.}
    \label{tab:dataset}
\end{table}

\begin{table*}[t]
    \centering
    \scriptsize
    \begin{tabular}{p{0.8cm} | C{0.8cm} C{0.8cm} C{0.8cm} | C{0.8cm} C{0.8cm} C{0.8cm} |  C{0.8cm} C{0.8cm} C{0.8cm} | C{0.8cm} C{0.8cm} C{0.8cm} }
    \ChangeRT{1pt}
    & \multicolumn{3}{c}{\textbf{Search Ads}} & \multicolumn{3}{|c}{\textbf{Quora}} & \multicolumn{3}{|c}{\textbf{News}} & \multicolumn{3}{|c}{\textbf{Wiki}} \\
    \cmidrule(lr){1-1}
    \cmidrule(lr){2-4}
    \cmidrule(lr){5-7}
    \cmidrule(lr){8-10}
    \cmidrule(lr){11-13}
        \textbf{Method} & 
        \textbf{R10} & \textbf{R50} & \textbf{R100} & \textbf{R10} & \textbf{R50} & \textbf{R100} & \textbf{R10} & \textbf{R50} & \textbf{R100} & \textbf{R10} & \textbf{R50} & \textbf{R100} \\
        \hline
        {MoPQ$^{b}$} & {0.2141} & {0.3513} & {0.4193} & {0.5861} & {0.8381} & {0.9054} & {0.3422} & {0.5462} & {0.6334} & {0.1005} & {0.2150} & {0.2878} \\   
        {MoPQ$^{a}$} & \textbf{0.2439} & \textbf{0.3820} & \textbf{0.4563} & \textbf{0.6717} & \textbf{0.8976} & \textbf{0.9416} & \textbf{0.3799} & \textbf{0.5787} & \textbf{0.6533} & \textbf{0.1434} & \textbf{0.2658} & \textbf{0.3322} \\
        \hline
        {PQ}   & 0.1252 & 0.2126 & 0.2682 & 0.3503 & 0.6353 & 0.7518 & 0.2028 & 0.4040 & 0.5069 & 0.0340 & 0.1015 & 0.1496 \\
        {OPQ}  & 0.1506 & 0.2507 & 0.3127 & 0.4572 & 0.7349 & 0.8359 & 0.2381 & 0.4277 & 0.5180 & 0.0558 & 0.1409 & 0.2082 \\
        \hline
        {DQN}  & 0.1433 & 0.2520 & 0.3145 & \underline{0.5436} & \underline{0.8285} & \underline{0.8977} & \underline{0.3138} & \underline{0.5217} & \underline{0.6108} & 0.0720 & 0.1728 & 0.2451 \\
        {DVSQ} & 0.1548 & 0.2784 & 0.3498 & 0.4061 & 0.7097 & 0.8179 & {0.3045} & {0.5105} & {0.6102} & \underline{0.0841} & \underline{0.1849} & \underline{0.2602} \\
        {SPQ}  & \underline{0.1907 }& \underline{0.3145} & \underline{0.3844} & {0.5321} & {0.7984} & {0.8782} & 0.2470 & 0.4474 & 0.5449 & 0.0655 & 0.1648 & 0.2306 \\
        {DPQ}  & 0.1709 & 0.2953 & 0.3650 & 0.4941 & 0.7760 & 0.8621 & 0.2552 & 0.4641 & 0.5639 & 0.0599 & 0.1561 & 0.2254 \\ 
        \ChangeRT{1pt}
    \end{tabular}
    \caption{\small Comparisons between MoPQ (upper), non-supervised (middle) and supervised (lower) PQ baselines.}
    \label{tab:exp-main}
\end{table*}

\begin{table}[t]
    \centering
    \scriptsize
    \begin{tabular}{p{1.0cm} C{1.0cm} C{1.0cm} C{1.0cm} C{1.0cm} }
    \ChangeRT{1pt}
      \textbf{Method} & \textbf{Ads} & \textbf{Quora} & \textbf{News} & \textbf{Wiki} \\
      \hline
      PQ   & 21.4227 & 13.6768 & 16.1511 & 16.9847 \\ 
      DQN  &  7.5604 &  5.6794 &  9.2987 &  6.9921 \\ 
      MoPQ$^{b}$ & 35.7987 & 29.7244 & 26.4847 & 26.5861 \\
      MoPQ$^{a}$ & 39.5576 & 30.0319 & 27.0710 & 26.8888 \\
    \ChangeRT{1pt}
    \end{tabular}
    \caption{\small Comparison of reconstruction loss.}
    \label{tab:exp-l2}
\end{table}

We implement two MoPQ alternatives: 1) \textbf{MoPQ}$^{b}$, the basic form with MCL and conventional in-batch sampling; 2) \textbf{MoPQ}$^a$, the advanced form with both MCL and DCS.


$\bullet$ \textbf{Implementations}. We use BERT-like Transformers \citep{devlin2018bert} for text encoding: the \#layer is 4 and the hidden-dimension is 768. The input text is uncased and tokenized with WordPiece \citep{wu2016google}. The algorithms are implemented in PyTorch 1.8.0. We consider the following codeword selection functions 
: 1) $l2$ (default option in reality), where the codeword is selected based on Euclidean distance: $\mathbf{C}_{ij} \leftarrow argmax \{-\|\mathbf{z}^{k}_i-\mathbf{C}_{i*}\|_2\}$; 2) Cosine, where the codeword is selected by: $\mathbf{C}_{ij} \leftarrow argmax \{cos(\mathbf{z}^{k}_i, \mathbf{C}_{i*})\}$; 3) Product, where the codeword is selected by: $\mathbf{C}_{ij} \leftarrow argmax \{\langle \mathbf{z}^{k}_i, \mathbf{C}_{i*} \rangle\}$; 3) Bi-linear, where the codeword is selected by: $\mathbf{C}_{ij} \leftarrow argmax \{\mathbf{z}^{k}_i W \mathbf{C}_{i*}^T\}$ ($W$ is a learnable square matrix). \emph{Our code and data will be made public-available. Pseudo codes for the algorithms, more comprehensive results and implementation details are put into supplementary materials.}


\begin{table*}[t]
    \centering
    \scriptsize
    \begin{tabular}{p{1.0cm} | C{0.8cm} C{0.8cm} C{0.8cm} | C{0.8cm} C{0.8cm} C{0.8cm} |  C{0.8cm} C{0.8cm} C{0.8cm} | C{0.8cm} C{0.8cm} C{0.8cm} }
    \ChangeRT{1pt}
    & \multicolumn{3}{c}{\textbf{Search Ads}} & \multicolumn{3}{|c}{\textbf{Quora}} & \multicolumn{3}{|c}{\textbf{News}} & \multicolumn{3}{|c}{\textbf{Wiki}} \\
    \cmidrule(lr){1-1}
    \cmidrule(lr){2-4}
    \cmidrule(lr){5-7}
    \cmidrule(lr){8-10}
    \cmidrule(lr){11-13}
        \textbf{Method} & 
        \textbf{R10} & \textbf{R50} & \textbf{R100} & \textbf{R10} & \textbf{R50} & \textbf{R100} & \textbf{R10} & \textbf{R50} & \textbf{R100} & \textbf{R10} & \textbf{R50} & \textbf{R100} \\
        \hline
        In-Batch & {0.2141} & {0.3513} & {0.4193} & {0.5861} & {0.8381} & {0.9054} & {0.3422} & {0.5462} & {0.6334} & {0.1005} & {0.2150} & {0.2878} \\
        {NCS} & 0.2249 & 0.3620 & 0.4305 & 0.6264 & 0.8655 & 0.9197 & 0.3569 & 0.5534 & 0.6355 & 0.1151 & 0.2320 & 0.2960 \\
        {DCS} & \textbf{0.2439} & \textbf{0.3820} & \textbf{0.4563} & \textbf{0.6717} & \textbf{0.8976} & \textbf{0.9416} & \textbf{0.3799} & \textbf{0.5787} & \textbf{0.6533} & \textbf{0.1434} & \textbf{0.2658} & \textbf{0.3322} \\
        \hline \hline
        {$l2$} & \textbf{0.2141} & \textbf{0.3513} & \textbf{0.4193} & 0.5861 & 0.8381 & \textbf{0.9054} & \textbf{0.3422} & \textbf{0.5462} & \textbf{0.6334} & 0.1005 & 0.2150 & 0.2878 \\
        {Cosine} & 0.2019 & 0.3297 & 0.3987 & 0.5804 & 0.8296 & 0.8993 & 0.3215 & 0.5160 & 0.6031 & 0.0784 & 0.1828 & 0.2518 \\
        {Product} & 0.2027 & 0.3290 & 0.3998 & 0.5578 & 0.8001 & 0.8707 & 0.3256 & 0.5307 & 0.6158 & 0.0843 & 0.1930 & 0.2617 \\
        {Bi-Lin} & 0.2102 & 0.3410 & 0.4113 & \textbf{0.5878} & \textbf{0.8382} & 0.9034 & 0.3197 & 0.5210 & 0.6116 & \textbf{0.1115} & \textbf{0.2259} & \textbf{0.2965} \\
    \ChangeRT{1pt}
    \end{tabular}
    \caption{\small The impact of DCS (upper table), and the impact of codeword selection functions (lower table).}
    \label{tab:exp-abl}
\end{table*}

\begin{table}[t]
    \centering
    \scriptsize
    \begin{tabular}{p{2.4cm} | C{0.8cm} C{0.8cm} C{0.8cm} C{0.8cm} }
    \ChangeRT{1pt}
      \textbf{Method} & \textbf{Ads} & \textbf{Quora} & \textbf{News} & \textbf{Wiki} \\
      \hline
      SPQ (M=4, L=128)  & 0.1018 & 0.2804 & 0.1569 & 0.0461 \\ 
      MoPQ$^{b}$ (M=4, L=128)  & \textbf{0.1068} & \textbf{0.4484} & \textbf{0.2327} & \textbf{0.0563} \\
      \hline
      SPQ (M=4, L=256)  & 0.1264 & 0.3006 & 0.1833 & 0.0440 \\ 
      MoPQ$^{b}$ (M=4, L=256)  & \textbf{0.1289} & \textbf{0.4648} & \textbf{0.2649} & \textbf{0.0569} \\ 
      \hline
      SPQ (M=8, L=256)  & 0.1907 & 0.5321 & 0.2470 & 0.0655 \\ 
      MoPQ$^{b}$ (M=8, L=256)  & \textbf{0.2141} & \textbf{0.5861} & \textbf{0.3422} & \textbf{0.1005} \\ 
      \hline
      SPQ (M=16, L=256) & 0.2196 & 0.5767 & 0.2780 & 0.1029 \\ 
      MoPQ$^{b}$ (M=16, L=256) & \textbf{0.2574} & \textbf{0.6395} & \textbf{0.3889} & \textbf{0.1457} \\ 
    \ChangeRT{1pt}
    \end{tabular}
    \caption{\small Impact of codebook size. (Recall@10)}
    \label{tab:exp-mp}
\end{table}

\subsection{Experiment Analysis}
The experimental studies focus on three major issues: 1) the overall comparisons between MoPQ and the existing PQ baselines; 2) the impact of DCS; 3) the impacts of codebook configurations, like codeword selection and codebook size. 

$\bullet$ \textbf{Overall Comparisons}. The overall comparison results are shown in Table \ref{tab:exp-main}. The matching accuracy is measured by \textbf{Recall@N} (R@N). We use 8 codebooks by default, each of which has 256 codewords ($M=8$, $L=256$). The best performances are marked in bold; the most competitive baseline performances are underlined.

Firstly, it can be observed that the basic form MoPQ$^{b}$ consistently outperforms all the baselines, with 7.8\%$\sim$19.5\% relative improvements over the most competitive baselines on different datasets. With the enhancement of DCS, MoPQ$^{a}$ further improves the performances by 11.0\%$\sim$42.7\%, relatively. Both observations validate the effectiveness of our proposed methods. As for the baselines: the supervised PQ's performances are comparatively higher than the non-supervised ones; however, the improvements are not consistent: in some cases, OPQ achieves comparable or even higher recall rates than some of the supervised PQ. 

Secondly, we further clarify the relationship between the reconstruction loss and the query-key matching accuracy. More results on reconstruction loss are reported in Table \ref{tab:exp-l2}. For one thing, we find that by jointly learning the embedding and quantization models, DQN's reconstruction losses become significantly lower than PQ. At the same time, DQN's recall rates are consistently higher than PQ. Such observations indicate that: {to some extent, the reduction of reconstruction loss may help to improve PQ's retrieval accuracy}. On the other hand, the reconstruction losses of MoPQ$^b$ and MoPQ$^a$ are much higher than DQN, but it dominates the baselines in terms of recall rate. Such observations echo the theoretical and empirical findings in Section \ref{sec:2-2}: {PQ's query-key retrieval accuracy will not monotonously increase with the reduction of reconstruction loss}.

A brief conclusion for the above observations: although the minimization of reconstruction loss still contributes to PQ's retrieval accuracy, the improvement is limited due to the non-monotone between both factors. In contrast, the minimization of MCL directly maximizes the query-key's matching probability, which makes MoPQ achieve much more competitive retrieval accuracy.

$\bullet$ \textbf{Impact of DCS}. More analysis about DCS is shown in the upper of Table \ref{tab:exp-abl}. We consider two baselines: 1) the conventional in-batch sampling, where no cross-device sampling is made; 2) the Non-differentiable Cross-device Sampling (NCS), which also makes embeddings shared across GPU devices for contrastive sample augmentation, but no image loss is computed to compensate the stopped gradients. It is observed that both DCS and NCS outperform the in-batch sampling, thanks to the augmentation of contrastive samples. However, the improvement of NCS is limited compared with DCS. As discussed in Section \ref{sec:pq-DCS}, the gradients are stopped at the non-differentiable variables of NCS, which causes distortions for the model's update. Thus, the model's training outcome is restricted because of it.

$\bullet$ \textbf{Impact of codeword selection}. We use MoPQ$^b$ as the representative to analyze the impact of different forms of codeword selections in the lower of Table \ref{tab:exp-abl}. We find that MoPQ's performances are not sensitive to the codeword selection function, as the experiment results are quite close to each other. Given that the $l$2 selection's performance is slightly better and no extra parameters are introduced, it is used as our default choice. 


$\bullet$ \textbf{Impact of codebook size}. We analyze the impact of codebook size in Table \ref{tab:exp-mp}; the SPQ baseline is included for comparison. With the expansion of scale, i.e., more codebooks and more codewords per codebook, MoPQ's performance can be improved gradually. In all of the settings, MoPQ maintains its advantage over SPQ. It is also observed that in some cases, MoPQ outperforms SPQ with even smaller codebook size, e.g., MoPQ (M=4, L=256) and SPQ (M=8, L=256) on News; in other words, the higher recall rate is achieved with smaller space and time costs. 


\section{Conclusion}
In this paper, we propose MoPQ to optimize PQ's ad-hoc retrieval accuracy. A systematic revisit is made for the existing supervised PQ, where we identify the limitation of using reconstruction loss minimization as the training objective. We propose MCL as our new training objective, where the model can be learned to maximize the query-key matching probability to achieve the optimal retrieval accuracy. We further leverage DCS for contrastive sample argumentation, which ensures the effective minimization of MCL. The experiment results on 4 real-world datasets validate the effectiveness of our proposed methods.


\bibliographystyle{acl_natbib}
\bibliography{anthology,acl2021}

\appendix
\section{Proof}

\textbf{Theorem 2.1.} (\textbf{Positive Reconstruction Loss}) The reconstruction loss is positive if the codebooks’ scale is smaller than the key’s scale.

\textbf{Proof.} The statement is proved by induction. First of all, it's trivial that the reconstruction loss is positive when the number of codewords is one ($L=1$). Suppose we are given codebooks $\mathbf{C}$, where $L>1$ and the corresponding reconstruction loss is positive. 
Now assgin a new codeword $\mathbf{C}_{*,L+1}$ to each codebook, where $[\mathbf{C}_{1,L+1},...,\mathbf{C}_{M,L+1}]=\mathbf{z}^k$ ($\mathbf{z}^k$ is an arbitrary key's embedding). With the augmentation of codebooks, the reconstruction loss related to $k$ will become zero, which means the reconstruction loss will be reduced by $\|\mathbf{z}^k-\tilde{\mathbf{z}}^k\|_2$. 
This is to say, when the size of codebooks is smaller than the number of keys, the reconstruction loss can always be reduced by increasing the codebooks' scale. Considering that the reconstruction loss is lower bounded by 0, it is obvious that the reconstruction loss is always positive given finite codebooks. $\square$


\textbf{Lemma 2.1.}
For each codebook $\mathbf{C}_i$, there will always exist perturbation vectors like $\epsilon_{i}$, where the manipulation of codewords: $\hat{\mathbf{C}}_{i*} \leftarrow \mathbf{C}_{i*}+\epsilon_{i}$, doesn't affect the codeword assignment.

\textbf{Proof.} 
The existence of $\epsilon_{i}$ is proved by 
giving one ``always-valid'' example.
We define the follow thresholding radius for the perturbation: 
\begin{gather*}\label{eq:radius}
\begin{split}
   r_{\epsilon_{i}} \gets & 0.5*min\{min\{||\mathbf{C}_{il}-\mathbf{z}_i||_2:l\neq j\}\\
   &-||\mathbf{C}_{ij}-\mathbf{z}_i||_2 : \forall \mathbf{z}\}, 
\end{split}
\end{gather*}
where $\mathbf{C}_{ij}$ is the assigned codeword for $i$-th sub-vector $\mathbf{z}_{i}$ of embedding $\mathbf{z}$, i.e., $j=argmin\{||\mathbf{C}_{i*}-\mathbf{z}_i||_2\}$.

Let $r \sim \mathrm{Uniform}(0,r_{\epsilon_{i}})$, and $\epsilon_{i}$$\gets$$r * \mathbf{u}_i$ ($\mathbf{u}_i$ is an arbitrary unit vector). In this way, $\forall l\neq j:$
\begin{equation}\label{ineq:epsilon}
    2||\epsilon_{i}||_2 < 2r_{\epsilon_{i}}\leq ||\mathbf{C}_{il}-\mathbf{z}_i||_2 - ||\mathbf{C}_{ij}-\mathbf{z}_i||_2
\end{equation}

Add this $\epsilon_i$ to all codewords in the codebook $\mathbf{C}_i$: $\hat{\mathbf{C}}_{i*} \leftarrow \mathbf{C}_{i*}+\epsilon_{i}$. According to the inequation (\ref{ineq:epsilon}) and triangle inequality, we can get:
\begin{equation*}
\begin{split}
    ||\hat{\mathbf{C}}_{ij}-\mathbf{z}_i||_2 
    \leq & ||\mathbf{C}_{ij}-\mathbf{z}_i||_2+||\epsilon_{i}||_2 \\
    < &||\mathbf{C}_{il}-\mathbf{z}_i||_2-||\epsilon_{i}||_2\\
    < &||\hat{\mathbf{C}}_{il}-\mathbf{z}_i||_2,
\end{split}
\end{equation*}
where $l\neq j$. Therefore, it's obvious that:
\begin{equation*}
j=argmin\{||\hat{\mathbf{C}}_{i*}-\mathbf{z}_i||_2\}.
\end{equation*}
The $\mathbf{z}_i$ will still be mapped to $j$-th codeword in codebook $\mathbf{C}_i$.
In other words, the original codeword assignment will not be affected by $\epsilon_i$. $\square$

\begin{table*}[t]
    \scriptsize
    \begin{tabular}
    {|c|l|ccc|ccc|ccc|ccc|}
    \hline
    & & \multicolumn{3}{c}{\textbf{Search Ads}} & \multicolumn{3}{|c}{\textbf{Quora}} & \multicolumn{3}{|c}{\textbf{News}} & \multicolumn{3}{|c|}{\textbf{Wiki}}  \\
    \hline
   \textbf{M\&N} & \textbf{Method} & 
    \textbf{R10} & \textbf{R50} & \textbf{R100} & \textbf{R10} & \textbf{R50} & \textbf{R100} & \textbf{R10} & \textbf{R50} & \textbf{R100} & \textbf{R10} & \textbf{R50} & \textbf{R100} \\
    \hline
    \multirowcell{8}{M=4\\N=128} & {PQ}   & 0.0248 & 0.0633 & 0.0895 & 0.1255 & 0.3214 & 0.4368 & 0.1455 & 0.3084 & 0.4002 & 0.0302 & 0.0872 & 0.1402 \\
    & {OPQ}  & 0.0365 & 0.0890 & 0.1252 & 0.2134 & 0.3214 & 0.4368 & 0.1388 & 0.3215 & 0.4218 & 0.0213 & 0.0676 & 0.1048 \\
    
    & {DQN}  & 0.0319 & 0.0846 & 0.1227 & 0.3618 & 0.6722 & 0.7759 & 0.0895 & 0.2250 & 0.3119 & 0.0241 & 0.0647 & 0.0975 \\
    & {DVSQ} & 0.0390 & 0.1005 & 0.1483 & 0.3855 & 0.7097 & 0.8118 & {0.1628} & {0.3460} & {0.4529} & {0.0413} & {0.0703} & {0.1383} \\
    
    & {SPQ}  & 0.1018 & 0.2166 & \underline{0.2882} & 0.2804 & 0.6095 & 0.7373 & 0.1569 & 0.3378 & 0.4424 & 0.0440 & 0.1098 & 0.1621 \\
    & {DPQ}  & 0.0454 & 0.1371 & 0.1980 & 0.2956 & 0.6431 & 0.7729 & 0.1434 & 0.3223 & 0.4260 & 0.0458 & 0.0927 & 0.1345\\
    & {MoPQ$^{b}$} & \underline{0.1068} & \underline{0.2199} & {0.2876} & \underline{0.4484} & \underline{0.7472} & \underline{0.8417} & \underline{0.2327} & \underline{0.4445} & \underline{0.5419} & \underline{0.0563} & \underline{0.1428} & \underline{0.2040} \\   
    & {MoPQ$^{a}$} & \textbf{0.1342} & \textbf{0.2567} & \textbf{0.3266} & \textbf{0.5130} & \textbf{0.8012} & \textbf{0.8815} & \textbf{0.2496} & \textbf{0.4447} & \textbf{0.5372} & \textbf{0.0742} & \textbf{0.1702} & \textbf{0.2341} \\
    \hline
    
    \multirowcell{8}{M=4\\N=256} & {PQ}   & 0.0348 & 0.0802 & 0.1388 & 0.1673 & 0.3893 & 0.5136 & 0.1455 & 0.3084 & 0.4002 & 0.0337 & 0.1071 & 0.1475 \\
    & {OPQ}  & 0.0533 & 0.1144 & 0.1581 & 0.2556 & 0.5253 & 0.6549 & 0.1640 & 0.3561 & 0.4634 & 0.0277 & 0.0851 & 0.1307 \\
    
    & {DQN}  & 0.0391 & 0.1040 & 0.1485 & 0.4421 & 0.7430 & 0.8341 & 0.1267 & 0.2892 & 0.3868 & 0.0320 & 0.0846 & 0.1238 \\
    & {DVSQ} & 0.0481 & 0.1287 & 0.1884 & 0.4103 & 0.7446 & 0.8446 & {0.2011} & {0.4080} & {0.5163} & {0.0447} & {0.1269} & {0.1842} \\
    
    & {SPQ}  & 0.1264 & {0.2416} & {0.3066} & {0.3006} & {0.6142} & {0.7360} & 0.1833 & 0.3754 & 0.4679 & 0.0461 & 0.1130 & 0.1635 \\
    & {DPQ}  & 0.0630 & 0.1627 & 0.2272 & 0.3111 & 0.6513 & 0.7688 & 0.1666 & 0.3606 & 0.4652 & 0.0493 & 0.0986 & 0.1446\\
    & {MoPQ$^{b}$} & \underline{0.1289} & \underline{0.2627} & \underline{0.3109} & \underline{0.4648} & \underline{0.7603} & \underline{0.8545} & \underline{0.2649} & \underline{0.4721} & \underline{0.5674} & \underline{0.0569} & \underline{0.1415} & \underline{0.2009} \\   
    & {MoPQ$^{a}$} & \textbf{0.1506} & \textbf{0.2802} & \textbf{0.3486} & \textbf{0.5285} & \textbf{0.8057} & \textbf{0.8817} & \textbf{0.2904} & \textbf{0.4909} & \textbf{0.5778} & \textbf{0.0854} & \textbf{0.1902} & \textbf{0.2615} \\
    \hline
    
    \multirowcell{8}{M=16\\N=256} & {PQ}   & 0.2184 & 0.3413 & 0.4123 & 0.4520 & 0.7870 & 0.8715 & 0.3093 & 0.5269 & 0.6194 & 0.0769 & 0.1993 & 0.2718 \\
    & {OPQ}  & 0.2260 & 0.3588 & 0.4227 & 0.3788 & 0.5854 & 0.6693 & 0.2381 & 0.4277 & 0.5180 & 0.1160 & 0.2322 & 0.3073 \\
    
    & {DQN}  & 0.2347 & 0.3718 & 0.4414 & 0.6061 & 0.8538 & 0.9100 & 0.3667 & 0.5629 & 0.6478 & 0.1411 & 0.2784 & 0.3533 \\
    & {DVSQ} & 0.2257 & 0.3489 & 0.4156 & 05819. & 0.8352 & 0.9046 & {0.3327} & {0.5420} & {0.6271} & {0.1291} & {0.2538} & {0.3245} \\
    
    & {SPQ}  & 0.2196 & {0.3574} & {0.4258} & {0.5767} & {0.8354} & {0.8943} & 0.2780 & 0.4867 & 0.5798 & 0.1029 & 0.2210 & 0.2964 \\
    & {DPQ}  & 0.2196 & 0.3541 & 0.4222 & 0.5254 & 0.7990 & 0.8781 & 0.3241 & 0.5260 & 0.6105 & 0.0952 & 0.2102 & 0.2845\\
    & {MoPQ$^{b}$} & \underline{0.2574} & \underline{0.3893} & \underline{0.4608} & \underline{0.6395} & \underline{0.8732} & \underline{0.9291} & \underline{0.3889} & \underline{0.5814} & \underline{0.6596} & \underline{0.1654} & \underline{0.2946} & \underline{0.3697} \\   
    & {MoPQ$^{a}$} & \textbf{0.2722} & \textbf{0.4176} & \textbf{0.4874} & \textbf{0.7075} & \textbf{0.9118} & \textbf{0.9506} & \textbf{0.4411} & \textbf{0.6217} & \textbf{0.6909} & \textbf{0.2065} & \textbf{0.3411} & \textbf{0.4077} \\
    \hline
    \end{tabular}
    \vspace{-5pt}
    \caption{\small 
    Performance for different codebooks' scale. The best performances are marked in bold, and the second are underlined.}
    \vspace{-10pt}
    \label{tab:exp-appendix}
\end{table*}

\section{Training Details}
The models are implemented with PyTorch 1.8.0 and run with 8$\times$Nvidia-A100-40G GPUs. 
We use an average pooling over the last transformers layer as the text embedding and then apply an additional linear layer to reduce the size of embedding to 128. The weight of reconstruction loss in baselines is tuned from \{1, 1e-1, 1e-2, 1e-3, 1e-4\}. 
Without explicit mention, we default use eight codebooks, each of which has 256 codewords.
We optimize the parameters with the Adam optimizer. The learning rate is 1e-4 for pretrained transformers layers and 1e-3 for other layers (e.g., product quantization). 
We train the models for 50 epochs with a batch size of 500.
We evaluate the model every epoch on the validation
set and keep the best checkpoint for the
final evaluation on test sets.

\section{More Experiments Results}
We conduct extensive experiments for MoPQ and baselines on four datasets. The results are shown in Table~\ref{tab:exp-appendix}.
The best performances with the same settings are marked in bold, and the second are underlined. The observations from this table are consistent with the conclusions in our paper: 
\begin{itemize}
    \item The performance of PQ models can be improved by increasing the number of codebooks and codewords. 
    \item MoPQ consistently  obtains better performance than baselines.
    \item Adopting DCS method to augment contrastive samples,  MoPQ$^{a}$ achieves significant improvement than MoPQ$^{b}$.
\end{itemize}



\end{document}